\documentclass[conference]{IEEEtran}
\IEEEoverridecommandlockouts
\usepackage{cite}
\usepackage{amsmath,amssymb,amsfonts}
\usepackage{algorithmic}
\usepackage{graphicx}
\usepackage{textcomp}
\usepackage{xcolor}
\usepackage{url}
\usepackage{balance}
\usepackage{multirow}
\usepackage{multicol}
\usepackage{hhline}
\usepackage{tabularray}
\usepackage{flushend}
\usepackage[numbers]{natbib}
\usepackage{graphicx}
\usepackage{booktabs}
\usepackage{diagbox}

\def\BibTeX{{\rm B\kern-.05em{\sc i\kern-.025em b}\kern-.08em
    T\kern-.1667em\lower.7ex\hbox{E}\kern-.125emX}}
\begin{document}

\title{Who Writes the Review, Human or AI?}
\author{
\IEEEauthorblockN{Panagiotis C. Theocharopoulos\IEEEauthorrefmark{1}, 
Spiros V. Georgakopoulos\IEEEauthorrefmark{2},
Sotiris K. Tasoulis\IEEEauthorrefmark{1} and
Vassilis P. Plagianakos\IEEEauthorrefmark{1}}
\IEEEauthorblockA{\IEEEauthorrefmark{1}Department of Computer Science and Biomedical Informatics}
\IEEEauthorblockA{\IEEEauthorrefmark{2}Department of Mathematics\\
University of Thessaly, Greece\\
Email: \{ptheochar, spirosgeorg, stasoulis, vpp\}@uth.gr}
}

\maketitle

\begin{abstract}
With the increasing use of Artificial Intelligence in Natural Language Processing, concerns have been raised regarding the detection of AI-generated text in various domains. This study aims to investigate this issue by proposing a methodology to accurately distinguish AI-generated and human-written book reviews. Our approach utilizes transfer learning, enabling the model to identify generated text across different topics while improving its ability to detect variations in writing style and vocabulary. To evaluate the effectiveness of the proposed methodology, we developed a dataset consisting of real book reviews and AI-generated reviews using the recently proposed Vicuna open-source language model. The experimental results demonstrate that it is feasible to detect the original source of text, achieving an accuracy rate of 96.86\%. Our efforts are oriented toward the exploration of the capabilities and limitations of Large Language Models in the context of text identification. Expanding our knowledge in these aspects will be valuable for effectively navigating similar models in the future and ensuring the integrity and authenticity of human-generated content.
\end{abstract}


\section{Introduction}
The introduction of the transformer architecture, along with the attention mechanism, triggered a paradigm shift in Natural Language Processing (NLP) research. Its unique capacity to simultaneously process full input sequences superseded Recurrent Neural Networks' previous dominance. This innovation showed the way for the widespread use of Large Language Models (LLMs), which have billions of parameters and are extensively trained on huge datasets. Furthermore, the introduction of pre-trained models enables language models to acquire complete knowledge of generic language patterns and structures from different and large data sources, allowing them to be refined for specific tasks~\cite{KASNECI2023102274}.

LLMs, which leverage massive datasets utilizing state-of-the-art Machine Learning techniques, have demonstrated an unprecedented ability to understand and generate human-like text responses. One of the most significant implications of LLMs is their capacity to revolutionize how we communicate with machines. Traditionally, interacting with computers has required the use of specific commands or syntax, which can be cumbersome and unintuitive for many users. LLMs, however, can understand and respond to natural language queries, making it easier for individuals to engage with technology in a more human-like manner. Moreover, LLMs have the potential to greatly enhance our access to information. As a result, LLMs can make it easier for individuals to navigate the ever-growing sea of digital information, and may even help to bridge the gap between different cultures and languages. In addition to these practical applications, LLMs are also poised to transform various creative industries by automating certain aspects of content creation and leading to new possibilities in storytelling and artistic expression.

Despite these promising developments, the rise of LLMs also raises important ethical and societal concerns. To this end, scientists and researchers have expressed both enthusiasm and anxiety over the introduction of ChatGPT~\cite{schulman2022chatgpt} and similar LLMs. Despite great advancements in language models, there are legitimate concerns about their misuse~\cite{jawahar2020automatic}, such as increased bias and discrimination, erosion of trust in digital media, job displacement, privacy concerns, over-reliance on AI, ethical decision-making, and concentration of power to those having the resources required to train those huge models.

Another potential threat is the increased risk for misinformation and fake news generation since LLMs have reached a level where they can produce exceptionally high-quality texts, opening up numerous possibilities for their malicious application, e.g.\ generating text that mimics human language for a wide range of tasks, including coding, lyrics, document completion, and question answering~\cite{sadasivan2023can}. As the growth of such models continues, we see an interconnection of LLMs with other models, e.g.\ the LLMs in combination with image generation models can generate images from textual descriptions, or even in combination with Text-to-Speech (TTS) models can generate spoken output based on textual inputs. The potential applications of these LLMs are limitless and also a challenging task to be confidently detected. 

Up until now, the detection of synthetic text is proving to be extremely demanding, due to the continuous development of sophisticated text generation techniques to the point where they can even deceive human readers~\cite{ippolito2019automatic}. Consequently, there is a growing need to differentiate between generated and genuine text. Recently, this necessity has become even more substantial~\cite{ippolito2019automatic,crothers2022machine}.

Towards this goal, the primary scope of this research is to examine the effectiveness of utilizing Machine Learning techniques in various text topics generated by LLMs. Furthermore, we aim to enhance our understanding of the process by which artificial text is generated by investigating the prediction task through text analytics. The subsequent sections of this paper are organized as follows: Section~\ref{rel_work} provides an overview of the existing literature related to this topic, Section~\ref{method} outlines the methodology employed in this study, while in Section~\ref{results} is presented the experimental analysis of the study. In Section~\ref{conclusion}, we present the concluding remarks and suggest potential avenues for future research exploration.

\section{Related Work}\label{rel_work}
Despite being a relatively new concept, the recognition of artificially generated text by LLMs has already garnered attention in the research community. Today, several applications or methods have been introduced as they try to catch and identify AI-written text. For example, OpenAI released an AI text classifier. The model is trying to distinguish AI-generated from human texts. It's worth noting that the model requires more than 1,000 tokens to produce reliable results. However, this revised version has demonstrated better performance compared to the previous version, which was based on GPT-2. In the validation set, the Area Under the Curve (AUC) score improved from 0.95 to 0.97, indicating enhanced accuracy. In the challenge set, the AUC score increased from 0.43 to 0.66. In terms of identifying AI-generated text, the classifier correctly identifies 26\% of text generated by AI as potentially AI-written (true positives). However, it also makes the error of incorrectly identifying 9\% of human-written text as AI-generated (false positives)~\cite{openai_ind}. Additionally, similar works have been introduced, such as DetectGPT, which detects machine-generated text from LLMs. DetectGPT is a more effective approach for detecting machine-generated text than existing zero-shot methods. DetectGPT can also accurately detect fake news articles~\cite{mitchell2023detectgpt}. 

A recent study~\cite{mitrovic2023chatgpt} explores the ability of machine learning models to distinguish between human-generated text and ChatGPT-generated text. The authors use a Transformer-based model approach to classify restaurant reviews as either generated by a human or by ChatGPT and compare its performance to a perplexity-based classification approach. The study tries to explain the model's performance in order to gain insights into the reasoning behind the model's decisions. The results show that their proposed approach performs with an accuracy of 79\%.

On the same page,~\cite{chen2023gpt} presents a new approach for detecting ChatGPT-generated and human-written text using language models. The authors collected and released a pre-processed dataset called OpenGPTText, which consists of rephrased content generated using ChatGPT. They then designed, implemented, and trained two different models for text classification, using the Robustly Optimized BERT Pretraining Approach (RoBERTa) and Text-to-Text Transfer Transformer (T5), respectively. The resulting model is referred to as GPT-Sentinel. The methodology involved fine-tuning approaches to distinguish human-written and ChatGPT-generated text, data collection from ChatGPT, establishing the OpenGPTText dataset, training the frozen RoBERTa with MLP, and fine-tuning the T5 model. The study reported an accuracy of 97\% on the test dataset.

In a previous investigation, we explored the challenges of distinguishing human written and artificially generated text using Large Language Models. The study examines the performance of different methodological approaches for detecting machine-generated text using the GPT-3 model to generate scientific paper abstracts of real research papers. The study tries to explore the capabilities of identifying machine-generated text. The findings suggest that a combination of a Word2Vec text representation method along with an LSTM model performed better in distinguishing real from AI-generated scientific abstracts than a transformer-based approach using BERT~\cite{theocharopoulos2023detection}.

\section{Research Methodology}\label{method}
\begin{figure*}[!ht]
    \centering
    \includegraphics[width=\linewidth]{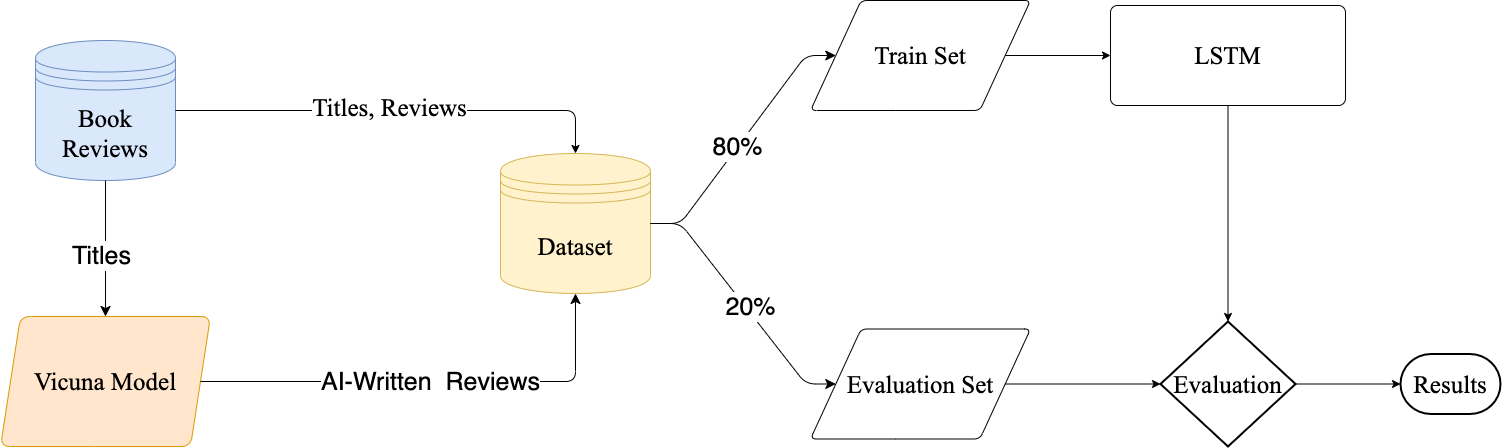}
    \caption{Schematic overview of this study. Titles and book reviews have been collected from the Kaggle Dataset. The titles of the selected books have been prompted to the Vicuna model, which returned the AI-written reviews based on their title. The study involved text cleaning and data representation as well as the model's results evaluation.}
    \label{overview}
\end{figure*}

The scope of this research is to develop a strategy that effectively discriminates human-written from AI-generated text. We aim to achieve generalization across domains by employing our pre-trained model on scientific abstracts~\cite{theocharopoulos2023detection}. Meanwhile, for the task at hand, we created a database of AI-generated book reviews by utilizing the open-source Vicuna model~\cite{vicuna2023}. The schematic overview of our method is illustrated in Figure~\ref{overview}.

\subsection{AI-Generated Dataset}
For this study, we used a comprehensive and publicly available book reviews dataset obtained from Kaggle~\cite{bekheet_2014_amazon}. The dataset consists of a substantial collection of reviews contributed by users on Amazon's Goodreads platform. It contains an extensive range of literary genres, with $3,000,000$ unique reviews across a diverse set of $10,883$ book categories.

The dataset's extensive coverage of different book categories and a large number of reviews allows our model to capture a wide range of language patterns and characteristics commonly found in human-written text. By training on this dataset, we aim to equip our model with the ability to recognize key differentiating factors between human and AI-generated text, enabling it to generalize and make accurate predictions on unseen data.

To generate AI-generated book reviews, we used the open-source Vicuna model, an open-source chatbot that was fine-tuned using user-shared conversations collected from ShareGPT~\cite{vicuna2023}. For creating the AI book reviews, we devised a suitable prompt in the following format: ``Act as a book critic. You have to write a short review regarding the book $\hat{t}$. Start the review with the text: [Review start] and end the review with the text [Review end]'', where $\hat{t}$ represents the original title of the book. To infuse creativity and novelty into the generated text, we made adjustments to the model's parameters. These adjustments included increasing the freedom to generate novelty by altering the randomness of the text generation process. The book titles used in the prompt were identical to those selected from the original dataset.

The resulting dataset consisted of $20,000$ entries, containing $10,000$ human-written reviews and $10,000$ AI-written reviews, all corresponding to the same book titles. To ensure the use of high-quality data in our analysis, a text-cleaning procedure was performed. This procedure involved several steps, including the removal of special characters or symbols, the elimination of whitespace and line breaks, the exclusion of stop words that do not convey meaningful information (e.g., "the", "and", "of"), the removal of non-alphabetic characters and numbers, normalization of the text by converting it to a consistent format (e.g., lowercase) to facilitate processing and analysis, the elimination of the response indicators of the LLM ``Review start'' and ``Review end'', and removal of the book titles from the reviews.

\subsection{Text representation}\label{text_representation}

Text representation is fundamental in the field of Natural Language 
Processing (NLP) as it transforms unstructured textual data into a structured and machine-readable format. The primary objective of text representation is to enable NLP algorithms to effectively handle various tasks, such as text classification, sentiment analysis, and machine translation.

One widely recognized technique for generating word embeddings, which are vector representations capturing both the semantic and syntactic meanings of words, is Word2Vec~\cite{mikolov2013efficient}. Word2Vec is a popular and extensively used NLP method that operates on the assumption that words appearing in similar linguistic contexts tend to possess similar semantic meanings. It encompasses two neural network-based models known as Continuous Bag of Words (CBOW) and Skip-gram, both of which are proficient in generating word embeddings from vast textual corpora. These embeddings effectively encode the semantic and syntactic relationships between words, thereby facilitating a wide range of NLP applications~\cite{mikolov2013efficient, goldberg2014word2vec}.

For our experimental purposes, we leveraged a pre-trained model called ``google-news-300'' which is an unsupervised NLP model~\cite{mikolov2013efficient}. This model, built upon the foundations of the Word2Vec technique, has been specifically designed to generate high-quality word embeddings from extensive collections of news articles. Each word within its vocabulary is represented by a 300-dimensional vector, therefore enabling the model to capture rich semantic information and subtle differences in contextual relationships.

\subsection{Model Architecture}
Artificial Neural Networks (ANNs) are computational models inspired by biological Neural Networks, designed for Machine Learning tasks~\cite{zhang2018deep}. ANNs consist of interconnected processing units called neurons, organized in layers, enabling parallel processing. Deep Neural Networks, a subclass of ANNs, have shown remarkable performance in solving challenging real-world problems. ANNs can be broadly classified into two types based on network topology: Feedforward Neural Networks (FNN) and Recurrent Neural Networks (RNNs). RNNs, unlike feedforward networks, utilize their internal state to process sequential inputs. One particular type of RNN is the Long Short-Term Memory (LSTM) network, which incorporates feedback connections to process entire data sequences. RNNs typically have a simple structure consisting of a chain of repeating modules. In contrast, LSTM networks feature four layers, including the hidden state and cell state, enhancing their modeling capabilities~\cite{zhang2018deep}. 

The training of the model is based on the well-known Transfer Learning (TL) technique. A popular technique for network training that transfers the knowledge from one problem domain to another (related) problem domain. This methodology is very useful in cases of datasets not adequate to train a full network due to small size, missing values, difficulties on annotations, etc.~\cite{georgakopoulos_TL,TL}. 

Leveraging the knowledge for detecting fake scientific abstracts~\cite{theocharopoulos2023detection}, we continue the training of the model on another text domain, the detection of AI-generated (fake) text on book reviews. The model's architecture used in this experiment contains an embedding layer. The embedding layer transforms the discrete word indices into dense vector representations. The pre-trained embeddings are not updated during the fine-tuning process, allowing the model to retain general knowledge. Next, an LSTM layer is added to capture the input data's sequential dependencies and temporal dynamics. In addition, to minimize the risk of overfitting and improve the model's generalization capabilities, a dropout layer is incorporated after the LSTM layer. The dropout regularization randomly drops out 50\% of the LSTM units during training, encouraging the network to learn more robust and diverse representations.
Finally, a Dense layer with a sigmoid activation function is added to produce the final output of the model.  For utilizing the transfer learning technique and fine-tuning the architecture, the pre-trained weights from the scientific abstracts classifier~\cite{theocharopoulos2023detection} have been loaded onto the model. The weights of the LSTM, Dropout, and Dense layers are updated during training, allowing the model to adapt to the target task and learn task-specific representations. 

\section{Experimental Results}\label{results}
In this section, we present the experimental results of our text classification model for distinguishing between human-written book reviews and AI-generated ones. We utilized a combination of the Word2Vec representation technique and LSTM networks. To train our model, we leveraged the weights of a previously trained classifier on COVID-19 scientific abstracts. To generate AI-written book reviews, we deployed the Vicuna model on a local machine equipped with an AMD EPYC 7502 CPU, 256GB RAM, and an Nvidia A100 80GB GPU.

Our analysis involved training and testing the model on a large corpus of original and AI-generated book reviews of interdisciplinary topics. The training set consisted of $80\%$ of the corpus, totaling $16,000$ entries, while the remaining $20\%$, comprising $4,000$ entries, was used for evaluation. It is important to note that each pair of real and generated pairs was exclusively present in either the train or test set. To assess the performance of our model, we utilized commonly used metrics such as accuracy, precision, recall, F1 score, and AUC. These metrics provide a comprehensive evaluation of the model's performance in terms of correctly classified instances, true positive rate, false positive rate, and the balance between precision and recall.

\begin{figure}[t]
    \centering
    \includegraphics[width=\linewidth]{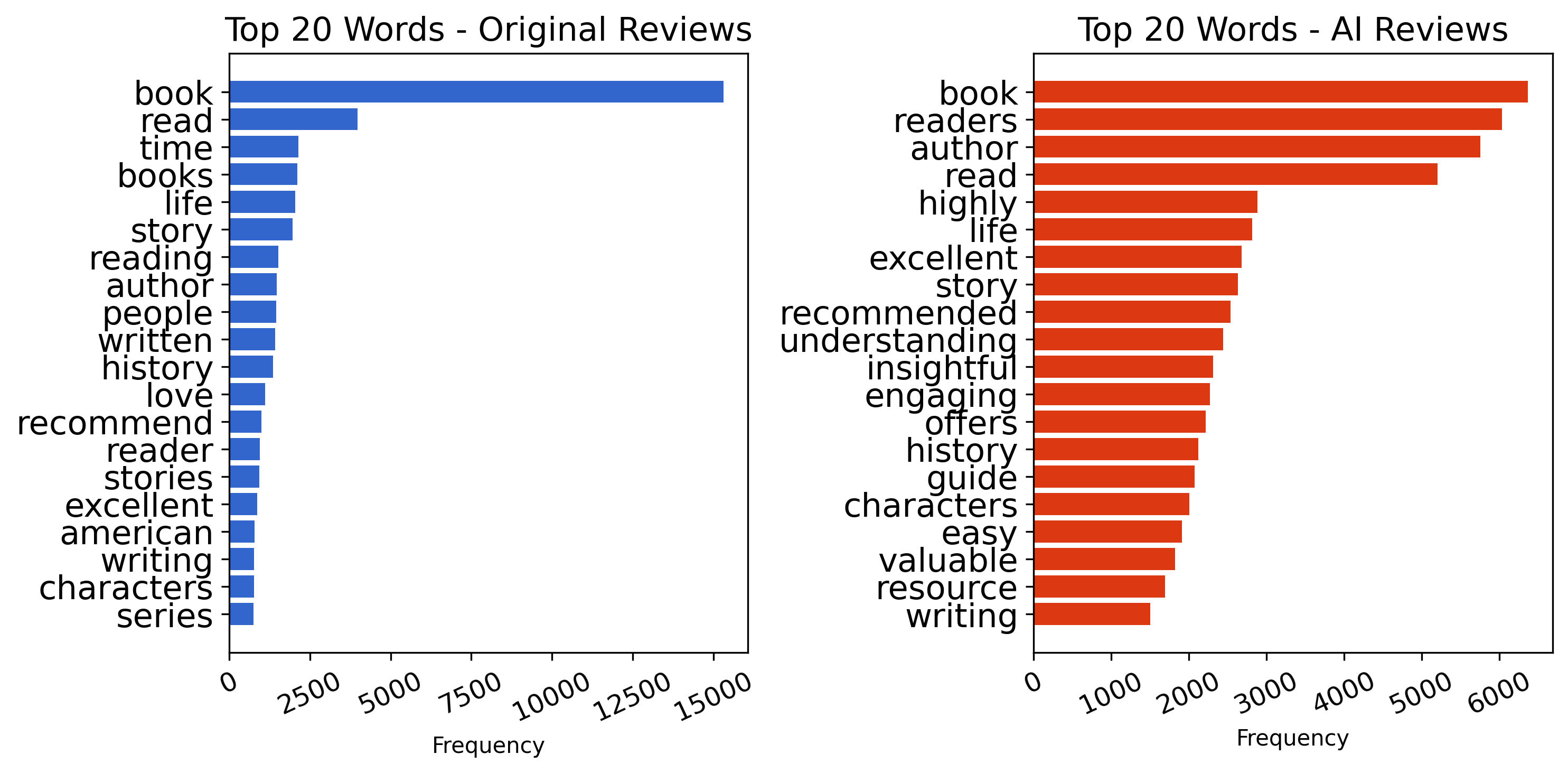}
    \caption{Frequencies of the top 20 words on both original and AI-generated reviews.}
    \label{word_freq}
\end{figure}

Before starting our model's training, we analyzed the reviews generated by the LLM. Figure~\ref{word_freq} visualizes the vocabulary shared between the original and AI-written reviews. It can be observed that both types of reviews utilize the same set of words. However, the frequency of occurrence of these words differs between human-written and AI-written reviews. Specifically, the frequency of words in human-written reviews tends to be higher compared to AI-generated reviews. 

In our preliminary results, we evaluate the proposed model without using the TL method, initializing the weights from the pre-trained model on detection of fake (AI-generated) scientific paper, but randomly initializing the weights of the model. Training the model without TL the classifier returned a mean accuracy of $0.924$. However, after implementing the TL method, the mean accuracy increased to $0.9686$. Table~\ref{res} displays the mean performance outcomes of the method across 100 separate iterations. To address the potential issue of overfitting, we monitored the validation loss. More specifically, as soon as the validation loss reached its minimum value and showed signs of subsequent increase, the training process was terminated and the current model was saved.


\begin{table}[htbp]
\centering
\caption{Average model evaluation scores with standard deviation after 100 independent runs before and after Transfer Learning.}
\label{res}
\begin{tabular}{lcc}
\hline
\textbf{Metrics} & \textbf{Mean Score - Non TL $\pm$ SD} & \textbf{Mean Score - TL $\pm$ SD} \\
\hline
Accuracy  & 0.92716 $\pm$ 0.019130 & 0.96866 $\pm$ 0.003209 \\
Precision & 0.92904 $\pm$ 0.048886 & 0.97300 $\pm$ 0.013589 \\
Recall    & 0.92910 $\pm$ 0.047390 & 0.96440 $\pm$ 0.017626 \\
F1-score  & 0.92730 $\pm$ 0.018455 & 0.96850 $\pm$ 0.003532 \\
AUC       & 0.92716 $\pm$ 0.018455 & 0.96866 $\pm$ 0.003209 \\
\hline
\end{tabular}
\end{table}

Furthermore, confidence intervals were calculated for each metric to estimate the range within which the true population parameter lies with a 95\% confidence level. As shown in Table~\ref{conf_inter}, these confidence intervals provide helpful insight into understanding the precision of the estimates.


\begin{table}[htbp]
\centering
\caption{Confidence Interval of model's evaluation scores.}
\label{conf_inter}
\begin{tabular}{lc}
\hline
\textbf{Metric} & \textbf{Confidence Interval} \\
\hline
Accuracy  & [0.9647, 0.9726] \\
Precision & [0.9561, 0.9899] \\
Recall    & [0.9425, 0.9863] \\
F1-score  & [0.9641, 0.9729] \\
AUC       & [0.9647, 0.9726] \\
\hline
\end{tabular}
\end{table}

\begin{figure}
    \centering
    \includegraphics[width=\linewidth]{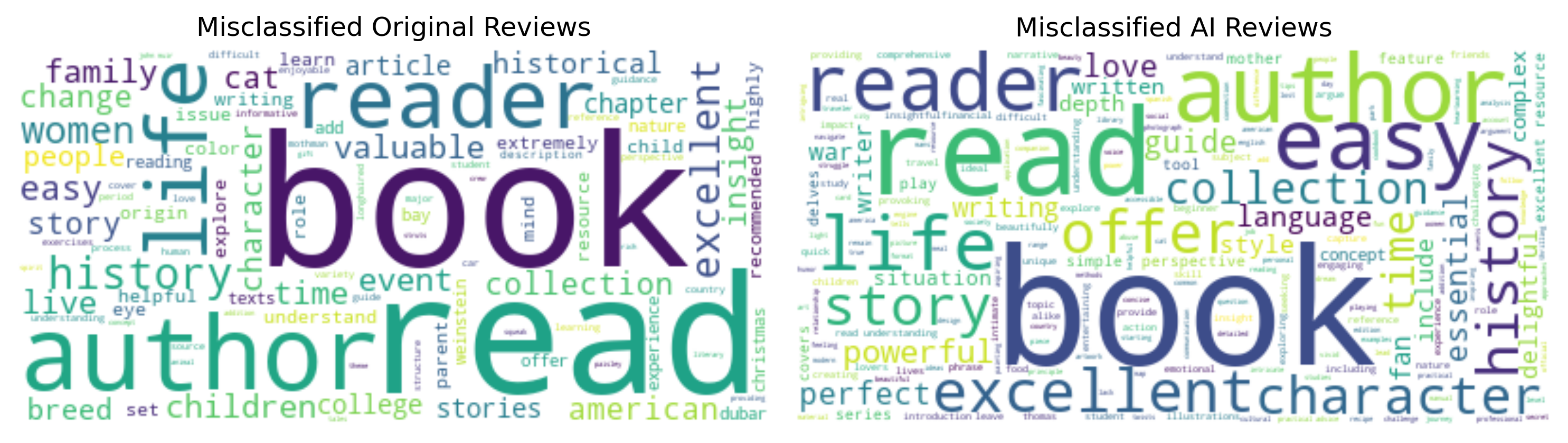}
    \caption{Wordcloud of the misclassified texts.}
    \label{wordcloud}
\end{figure}

\begin{figure}
    \centering
    \includegraphics[width=\linewidth]{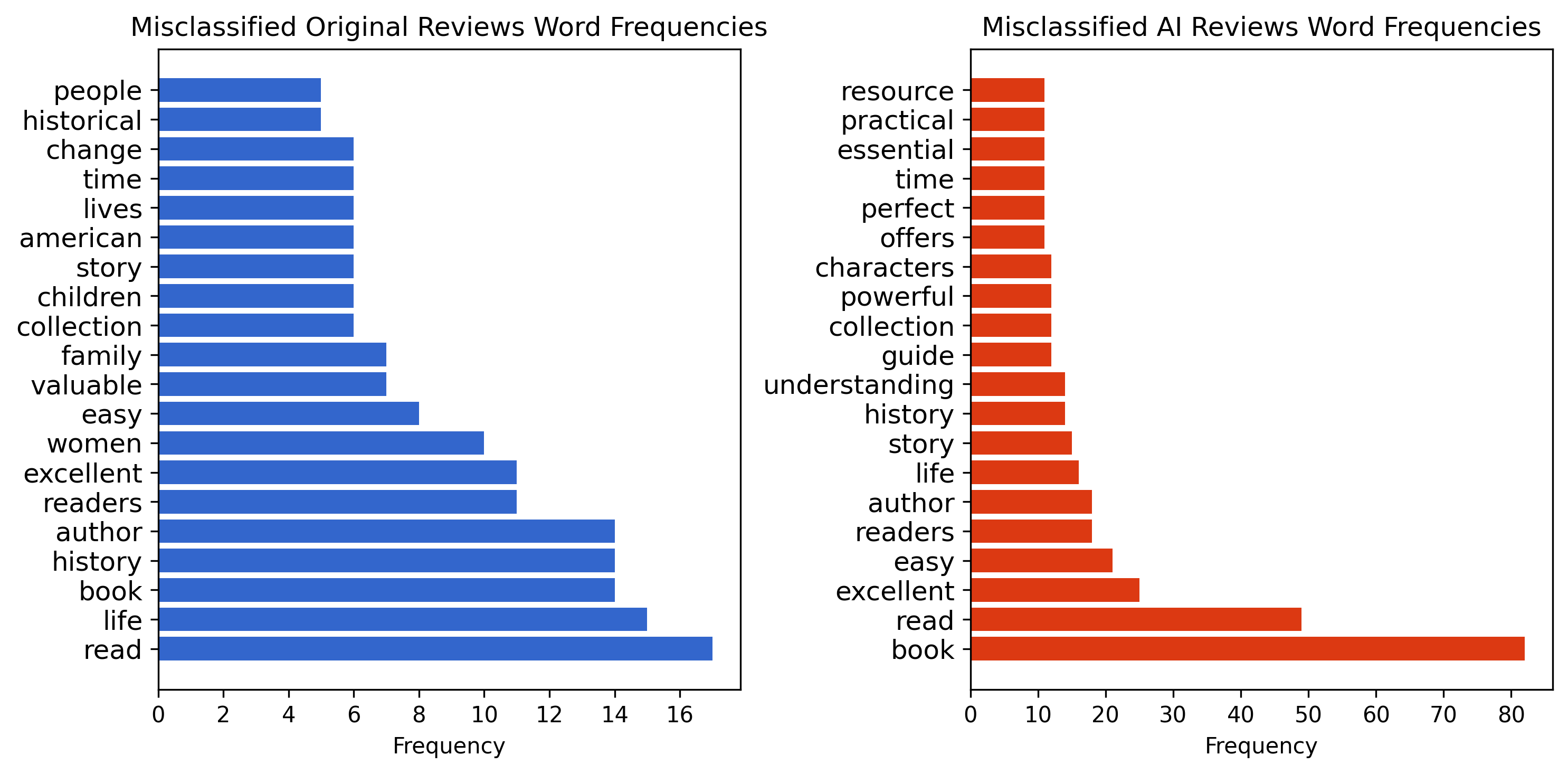}
    \caption{Frequencies of the misclassified words.}
    \label{miss_word_freq}
\end{figure}

If we now turn our attention to the misclassified results, we found that, on average, the model incorrectly classified 60 texts as AI-written instead of human-written. Conversely, it mislabeled 117 texts as human-written instead of AI-written. To gain further insights into the errors made by the model, we visualized the word usage in both cases using a wordcloud, as depicted in Figure~\ref{wordcloud}. This visualization provides a clearer understanding of the specific words that contributed to the misclassifications. As we notice, both texts contain similar vocabulary. Thus, Figure~\ref{miss_word_freq} presents the frequency of those words in both categories only on the misclassified texts. As we can see, the frequency of the words is high in both states.

\begin{figure}[htb!]
    \centering
    \includegraphics[width=\linewidth]{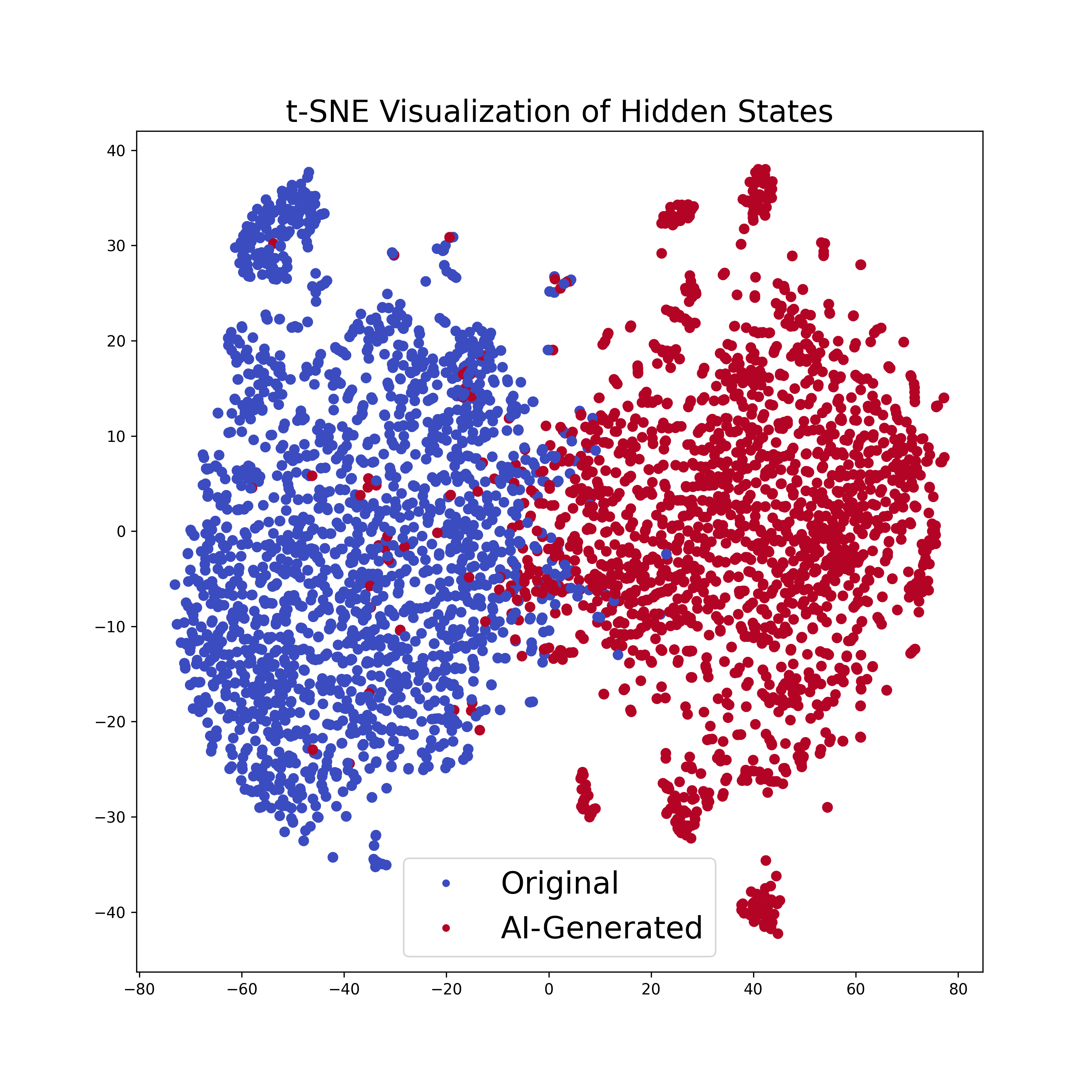}
    \caption{t-SNE Visualization of Hidden State Representation on the evaluation data. The blue points represent the original reviews and the red are the AI-generated text.}
    \label{tsne}
\end{figure}

To fully explore the performance of the model, in Fig.~\ref{tsne} we provide a visual representation of the evaluation data representations in the hidden state of our model. Using the t-SNE method, we reduced the dimensionality of the hidden states into a 2D space. The arrangement of points indicates how the model separates human-written and AI-Generated sequences. Notice that there are sequences that overlap, which was expected from the aforementioned results.

\section{Conclusion}\label{conclusion}

In this work, we focused on studying the feasibility of discrimination between text produced by Artificial Intelligence systems and human-generated text. In particular, we developed a methodology for this task that incorporates transfer learning, allowing us to fine-tune the model and enhance its ability to discern between real and AI-generated text across a wide range of topics. The proposed model achieved an average accuracy of $96.86\%$ in identifying the original and AI-written texts. Following the evaluation, we conducted an analysis of the misclassified texts. These sequences showed similarities with only minor variations in the choice of words. These findings provide valuable insights into the challenges of distinguishing between AI-generated and human-authored text.

In the future, we intend to examine the utilization of our proposed pre-trained model to other text domains and types, such as tweets, assuming that even without further training it will be able to detect the AI-generated texts.
Also, we intend to use novel, more advanced open-source language models to produce a larger dataset. This will allow us to refine our text-generating approach and test its efficacy across diverse domains, languages, and text types. We eventually hope to gain a deeper understanding of the potential and limitations of AI-produced content by studying the generated text.

\section*{Acknowledgment}
We acknowledge the support of this work by the project “Par-ICT CENG: Enhancing ICT research infrastructure in Central Greece to enable processing of Big data from sensor stream, multimedia content, and complex mathematical modeling and simulations” (MIS 5047244), which is implemented under the Action “Reinforcement of the Research and Innovation Infrastructure”, funded by the Operational Programme ”Competitiveness, Entrepreneurship and Innovation” (NSRF 2014-2020) and co-financed by Greece and the European Union (European Regional Development Fund).

\bibliographystyle{IEEEtranN}
\bibliography{bibliography}

\end{document}